\documentclass[10pt,twocolumn,letterpaper]{article}

\usepackage{cvpr}              
\definecolor{cvprblue}{rgb}{0.21,0.49,0.74}
\usepackage[pagebackref,breaklinks,colorlinks,allcolors=cvprblue]{hyperref}
\usepackage{caption}
\usepackage{multicol}
\usepackage{multirow}
\usepackage{pifont}
\usepackage{xcolor}
\usepackage{stackengine}
\usepackage{tablefootnote}
\usepackage{array}

\newcolumntype{C}[1]{>{\centering\arraybackslash}p{#1}}

\definecolor{myblue}{RGB}{39, 92, 173} 
\definecolor{mygreen}{RGB}{38, 140, 75}
\newcommand{\myBlue}[1]{\textcolor{myblue}{#1}}
\newcommand{\myGreen}[1]{\textcolor{mygreen}{#1}}

\newcommand{\cmark}{\textcolor{green!60!black}{\ding{51}}}
\newcommand{\xmark}{\textcolor{red}{\ding{55}}}

\def\paperID{} 
\def\confName{CVPR}
\def\confYear{2026}

\title{OnlineHMR: Video-based Online World-Grounded Human Mesh Recovery}

\author{Yiwen Zhao$^1$,
        Ce Zheng$^{1,\dagger,\ddagger}$,  
        Yufu Wang$^2$,
        Hsueh-Han Daniel Yang$^1$,
        Liting Wen$^1$,  
        L{\'a}szl{\'o} A. Jeni$^{1,\dagger}$ \\
 $^1$Carnegie Mellon University \quad $^2$University of Pennsylvania \\
\textsuperscript{\rm $\dagger$}Corresponding Authors, \textsuperscript{\rm $\ddagger$}Project Lead \\
}

\begin{document}

\twocolumn[{%
    \renewcommand\twocolumn[1][]{#1}
    \maketitle
    \vspace{-8mm}
    \begin{center}
        \centering
        \includegraphics[width=\textwidth]{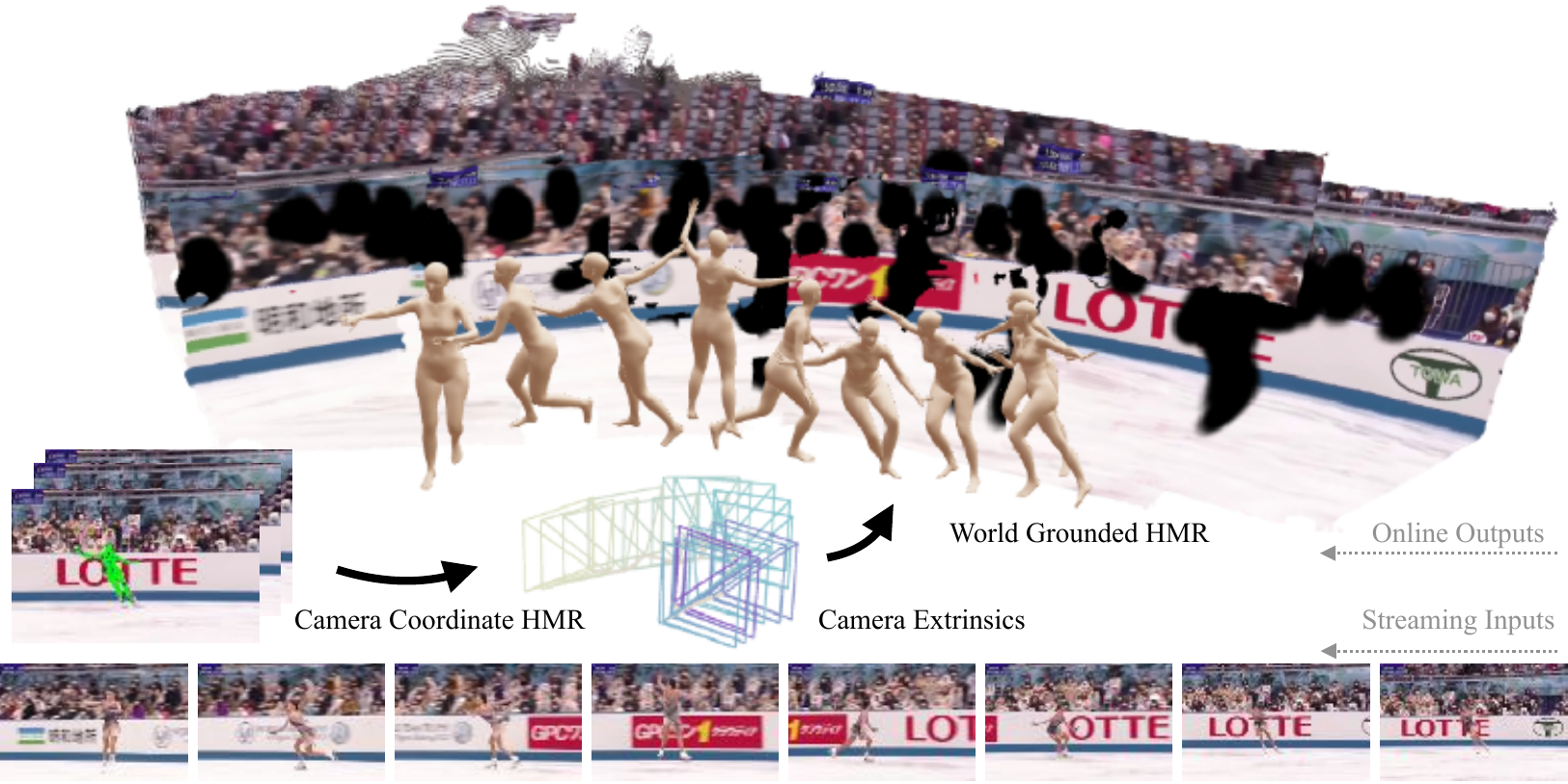}

        \captionof{figure}{An in-the-wild example of our framework. Given a streaming monocular RGB video, our method leverages a two-branch inference to recover the world-grounded human motion in an online manner.}
        \label{fig_teaser}
    \vspace{-1mm}
    \end{center}
}]

\maketitle
\begin{abstract}
Human mesh recovery (HMR) models 3D human body from monocular videos, with recent works extending it to world-coordinate human trajectory and motion reconstruction. However, most existing methods remain offline, relying on future frames or global optimization, which limits their applicability in interactive feedback and perception-action loop scenarios such as AR/VR and telepresence. To address this, we propose OnlineHMR, a fully online framework that jointly satisfies four essential criteria of online processing, including \textbf{system-level causality}, \textbf{faithfulness}, \textbf{temporal consistency}, and \textbf{efficiency}. Built upon a two-branch architecture, OnlineHMR enables streaming inference via a causal key–value cache design and a curated sliding-window learning strategy. Meanwhile, a human-centric incremental SLAM provides online world-grounded alignment under physically plausible trajectory correction. Experimental results show that our method achieves performance comparable to existing chunk-based approaches on the standard EMDB benchmark and highly dynamic custom videos, while uniquely supporting online processing. Page and code are available at \url{https://tsukasane.github.io/Video-OnlineHMR/}.
\end{abstract}    
\vspace{-5mm}

\section{Introduction}
\label{sec:intro}

Human mesh recovery (HMR) aims to reconstruct the 3D human body pose and shape from RGB images or videos. With the development of parametric body models~\cite{SMPL, smplx, park2025atlas}, HMR has become a core task in human-centric perception, supporting applications such as motion analysis~\cite{facing, gait}, motion retargeting for humanoids~\cite{luo2023perpetual, he2025asap}, and virtual reality~\cite{han2022umetrack, peng2022ase}. While early studies estimate human meshes in camera coordinates, recent works have advanced toward global human trajectory and movement recovery. This paradigm captures absolute translation, orientation, and scale in a world that is consistent along time, enabling a holistic understanding of human dynamics. 

Although global human trajectory and motion recovery approaches have made notable progress, most existing methods are still offline, requiring access to the entire video clip before inference. Such pipelines are impractical for real-time or interactive applications. In contrast, online processing, i.e., predicting the mesh output from only past and present inputs, offers low latency and immediate response. It also avoids the heavy computation and memory cost of long-sequence offline processing. Hence, enabling global human trajectory and motion recovery in an online manner is a critical step toward practical, real-world deployment.

To achieve online global human trajectory and motion recovery, a system must satisfy four essential conditions:
\begin{itemize}
  \item \textbf{System-level causality.} The entire pipeline, including both the camera-coordinate HMR and global trajectory estimation modules, must operate causally and incrementally. The prediction for frame $i$ should rely only on current and past inputs without accessing future frames or offline global optimization.
  \item \textbf{Faithfulness:} The reconstructed human meshes should accurately capture body geometry and pose details in each frame, ensuring reliable motion trajectories and physically plausible spatial alignment.
  \item \textbf{Temporal consistency:} Maintain smooth and coherent human motion over time, preserving geometric--kinematic continuity while avoiding long-term drift and excessive damping.
  \item \textbf{Efficiency:} Process each new frame with constant-time complexity, supporting continuous streaming inference with minimal latency.
\end{itemize}

Previous video-based HMR methods such as TCMR \cite{choi2021tcmr}, MPS-Net \cite{MPS-Net}, and GLoT~\cite{shen2023glot}
follow a similar design that takes a fixed-length 16-frame sequence as input and estimates the mesh of the \emph{center frame}. This strategy requires access to future frames and therefore violates the causal constraint. Moreover, this leads to redundant computation due to overlapping frames, resulting in low efficiency for continuous inference. Although WHAM~\cite{wham} claims to support online inference, in practice, only its \emph{camera-coordinate HMR} module achieves this goal. Its global trajectory estimation module still depends on an offline framework such as DPVO \cite{DPVO} or DROID-SLAM \cite{DROID-SLAM}, leveraging future frames to correct previous camera poses (details in Sec.~\ref{sec:exp4.2}).
Human3R \cite{chen2025human3r} supports online inference with a lightweight recurrent state, but it still falls short in accurately reconstructing humans and maintaining smooth and coherent motion over time. 

Thus, we propose \textbf{OnlineHMR}, a fully online framework for global human trajectory and motion recovery that jointly satisfies the four criteria discussed above. 
Built upon prior TRAM~\cite{tram} framework, OnlineHMR extends it to support streaming inference and causal processing while preserving high-fidelity reconstruction. 
To ensure \textbf{causality}, both the camera-coordinate HMR and the human-centric SLAM modules operate in a strictly incremental manner, where each frame prediction depends only on current and past observations. 
To enhance \textbf{faithfulness} under online constraint, we adopt a sliding-window learning strategy that leverages short-term temporal correlation during training while maintaining causal inference at test time, leading to accurate and stable mesh reconstruction. 
To guarantee \textbf{temporal consistency}, we introduce a velocity regularization loss in the camera-coordinate HMR branch to suppress abrupt motion changes, and employ an EMA correction for camera extrinsic to mitigate accumulated trajectory noise. 
Finally, OnlineHMR achieves high \textbf{efficiency} through a key–value cache inference design, enabling constant-time per-frame processing, while the human-centric SLAM performs incremental estimation without time-consuming global optimization. Together, these components enable OnlineHMR to perform reliable human motion recovery in world coordinates. 
\section{Related Works}
\label{sec:relatedworks}

\begin{figure*}[t]
  \centering
  \includegraphics[width=1.0\linewidth]
  {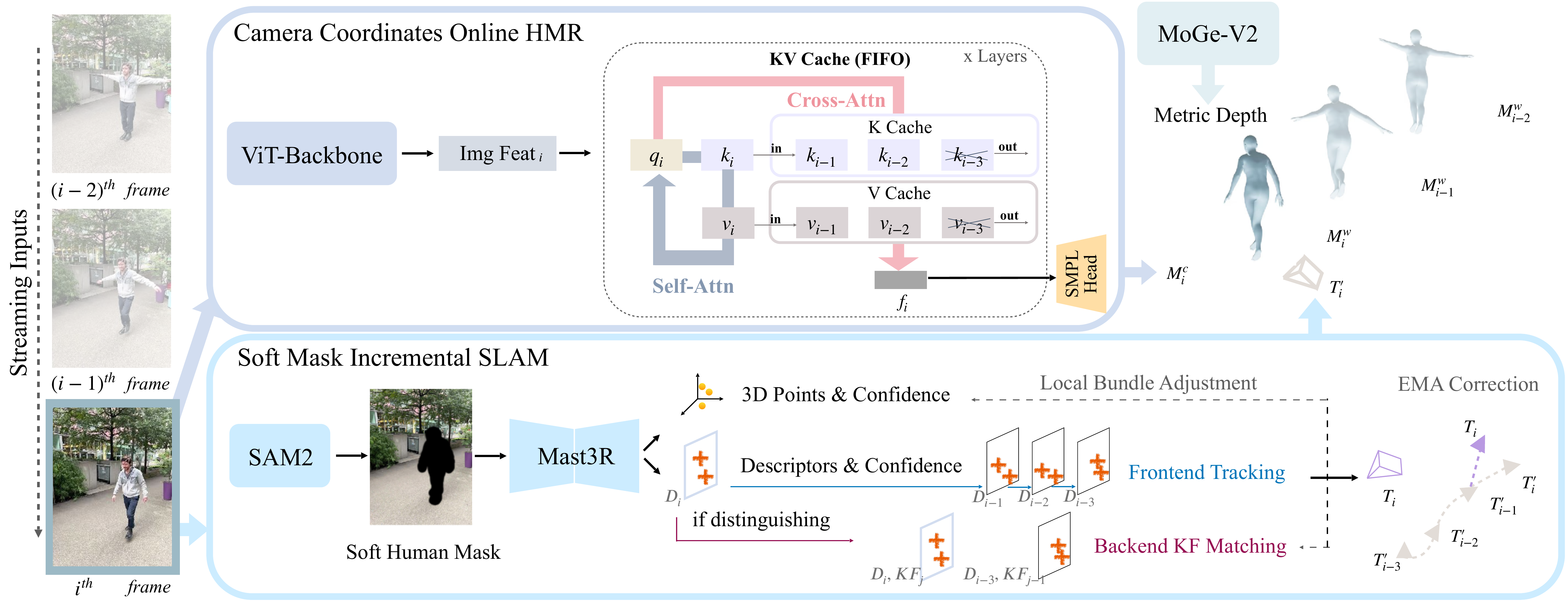}
  \caption{The online processing workflow of our method. Given a streaming video input, we estimate the world coordinate 3D human body of the most recent frame. $\mathbf{T}_i$ here is the homogeneous transformation matirx composed by $\mathbf{q}_i^{\text{c}}$ and $\mathbf{t}_i^{\text{c}}$ in Sec.~\ref{sec:method3.1}.}
  \label{fig_pipeline}
  \vspace{-4mm}
\end{figure*}

\subsection{Image and Video-based Human Mesh Recovery}
Image-based methods~\cite{hybrik2021Li,hmr2,refit} aim to estimate 3D human pose and shape from frame-wise spatial information. With kinematics decomposition~\cite{hybrik2021Li}, transformer adaptation~\cite{hmr2}, and well-designed feedback mechanism~\cite{refit}, these methods yield reasonably faithful result. More recent works paid attention to camera modeling, dense supervision, and pose prior. CameraHMR~\cite{camerahmr2025patel} trains a camera intrinsic predictor on human-centric images and incorporates dense 2D supervision. TokenHMR~\cite{tokenhmr2024dwivedi} leverages tokenization to define a valid pose prior space and introduces a novel TALS loss to alleviate the one-to-many mapping from 2D to 3D. NeuHMR~\cite{neuhmr2025xiang} uses neural rendered humans as dense 2D supervision. However, independently estimating each frame causes inconsistency in human scale and motion continuity. Video-based approaches~\cite{choi2021tcmr, MPS-Net, shen2023glot, yuan2022glamr, ye2023slahmr, wham} leverage cross-frame information to achieve temporally consistent 4D HMR, with focus switching from camera space to global space. 
Despite the rapid progress, for world-grounded HMR, the offline optimization methods~\cite{hsfm, ye2023slahmr, josh} take a long time to converge, while estimating global dynamics of humans~\cite{glamr} suffering from entanglement with camera motion. Recent works try to estimate the camera trajectory by VO~\cite{wham, shen2024gvhmr} or SLAM~\cite{tram}, but VO tends to underestimate the translation, and the use of globally optimized SLAM extrinsic is not online. Most relevantly, a concurrent work Human3R~\cite{chen2025human3r} provides a learning-based feedforward framework that  estimate scene reconstruction, human meshes, and camera trajectories in an online manner. However, a lack of details and accuracy in local human motion still exists. Based on these observations, we decouple the world trajectory and local motion estimation to expert models for faithfulness and temporal consistency, curating an online inference pipeline.

\subsection{Human-Camera-Scene Joint Modeling}
Human motion is entangled with camera motion as the extrinsic of the sensor influences global translation, orientation, and the intrinsic affects projection. It is also entangled with the environment in terms of contact and force. Faithful modeling of human motion requires a comprehensive consideration of these entities. Offline methods use optimization~\cite{josh, place, hsfm} or incorporate the simulation environment~\cite{videomimic, wang2025crisp} to place humans within proper scene geometry. Concurrent online method Human3R~\cite{chen2025human3r} builds upon the 4D scene-camera pretrained foundation model CUT3R\cite{cut3r} to further reconstruct humans, enabling a learned implicit constraint. But due to the limited data~\cite{bedlam} of humans compared to the 4D scene, the reconstructed world human still has unnatural jittering and inaccurate local motion. To address this, we design a two-branch framework, combining the camera and scene from SLAM with the human from camera coordinate HMR, while leveraging explicit constraints to minimize jitter and maintain trajectory faithfulness. 

\subsection{Streaming Inference}
Online processing targets videos that come in a streaming format, where each frame must be processed using only current and past information without access to future frames. Recent work demonstrates sophisticated approaches for maintaining temporal consistency under these conditions. StreamVGGT\cite{streamvggt} employs temporal causal attention where each token attends only to current and previous frames, and distills knowledge from bidirectional teacher models. Evict3R\cite{evict3r} addresses memory limitations in streaming transformers through training-free token eviction based on cumulative attention scores. CTVIS\cite{ctvis} and SAM 2\cite{sam2} address train-inference discrepancy through streaming memory banks and momentum-averaged embeddings. These streaming techniques share the insights of using history to provide clues for current estimation, inspiring our OnlineHMR design, where we leverage memory cache and physical constraints for temporal consistency.
\section{Methodology}

Given a streaming video, our goal is to reconstruct the SMPL~\cite{SMPL} parametric human model online in world coordinates. We first recap the definition of the SMPL model and the transformations between global and local coordinates in Sec.~\ref{sec:method3.1}. Next, we describe how to adapt a video-based HMR model for online operation in Sec.~\ref{sec:method3.2}. We then analyze the human–camera relationship and the associated constraints and their integration into an incremental SLAM system in Sec.~\ref{sec:method3.3}. Finally, we introduce a novel metric for evaluating the naturalness of the reconstructed motion in Sec.~\ref{sec:method3.4}.

\subsection{Parametric Human Model Preliminary}
\label{sec:method3.1}
Video-based HMR aims to reconstruct the 3D human body in a continuous time sequence. Following prior work~\cite{tram,wham}, we adopt the SMPL parametric model~\cite{SMPL}, which represents a human mesh $\mathbf{M}_i \in \mathbb{R}^{6890\times3}$ on frame index $i$ as a differentiable function of pose and shape parameters:
\begin{equation}
    \mathbf{M}_i = \mathcal{M}(\boldsymbol{\beta}_i, \boldsymbol{\theta}_i),
\end{equation}
where $\boldsymbol{\theta}_i \in \mathbb{R}^{23\times3}$ denotes the axis-angle rotations of body joints, and $\boldsymbol{\beta}_i \in \mathbb{R}^{10}$ encodes the body shape.
In addition, a rotation $\mathbf{R}_i^{\text{root}} \in \mathbb{R}^{3\times3}$ and translation $\mathbf{t}_i^{\text{root}} \in \mathbb{R}^{3}$ are estimated to position the body relative to the camera.
The HMR network predicts these parameters from visual inputs, producing the camera-coordinate mesh $\mathbf{M}_i^{\text{c}}$ that captures the human motion in the camera frame.

To recover the global human trajectory and motion in the world coordinate system, we transform the estimated camera-coordinate mesh $\mathbf{M}_i^{\text{c}}$ into the world space through a rigid transformation matrix:
\begin{equation}
    \mathbf{P}_i = [\mathbf{q}_i^{\text{c}}, s\cdot \mathbf{t}_i^{\text{c}}],
\end{equation}
where $\mathbf{q}_i^{\text{c}} \in \mathbb{R}^{4}$ and $\mathbf{t}_i^{\text{c}} \in \mathbb{R}^{3}$ represent the unit quaternion rotation and translation of the camera pose in an arbitrary scales. $s$ is a factor that converts the translation to metric scale.
The world-coordinate human mesh is obtained as
\begin{equation}
    \mathbf{M}_i^{\text{w}} = \mathbf{R(\mathbf{q}_i^{\text{c}})} \cdot \mathbf{M}_i^{\text{c}} + s \cdot \mathbf{t}_i^{\text{c}}.
\end{equation}
where $\mathbf{R(\mathbf{q}_i^{\text{c}})}$ denotes the rotation matrix converted from the unit quaternion. In our practice, we leverage an SLAM-based approach~\cite{MASt3r-slam} to estimate the camera trajectory $\{\mathbf{q}_i^\text{c}, \mathbf{t}_i^\text{c}\}$ across frames, and a metric depth estimator for the scaler $s$. Our framework processes both the camera trajectory and the camera coordinate human mesh in an online manner, while leveraging temporal clues and physical constraints from the video. We first describe the online recovery of the human mesh in camera coordinate system.

\subsection{Camera Coordinates Online HMR}
\label{sec:method3.2}
\noindent\textbf{Sliding Window Learning.} 
For camera-coordinate HMR, the key is to leverage temporal information from nearby frames to produce temporally smooth human motion in the current frame, which is challenging under the constraint of online processing. To address this, we divide the input sequence into short overlapping windows with a step size of 1 and perform feature fusion within each window using attention~\cite{attention}. Following TRAM~\cite{tram}, we initialize our model with the large-scale pretrained HMR2.0~\cite{hmr2}. Video frames are first split into patches, processed by a ViT~\cite{dosovitskiy2020vit} backbone, and converted into patch-level spatial features.

\begin{figure}[t]
    \centering    \includegraphics[width=1.0\linewidth]{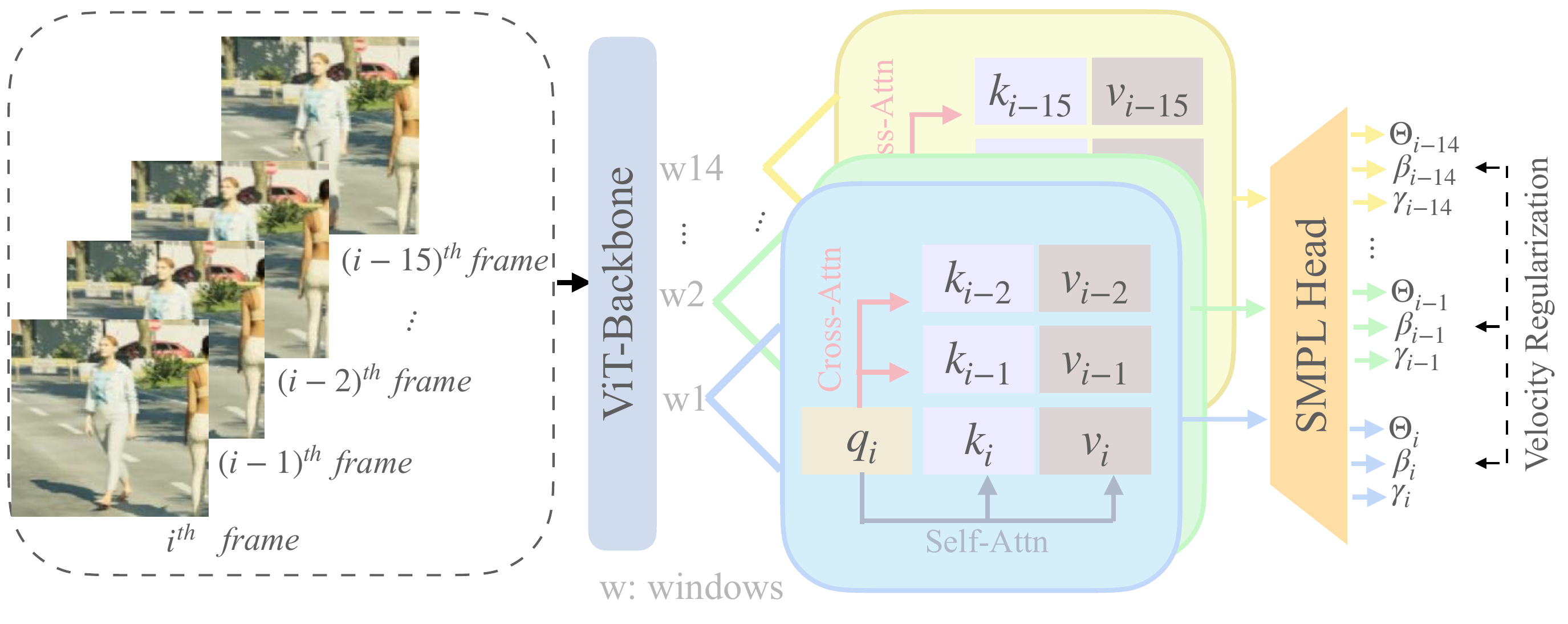}
    \caption{Sliding window learning pipeline. The input sequence is sliced to overlapping windows, learning spatial and temporal information fusion inside each window, and alleviate jitter effect through velocity regularization.}
    \vspace{-5mm}
    \label{fig_training}
\end{figure}

\noindent\textbf{\textit{Intra-window information fusion.}} Each temporal window contains $N$ frames, ranging from frame $i-N+1$ to frame $i$. As illustrated in Fig.~\ref{fig_training}, the last frame in each window performs self-attention on its own features while attending to preceding frames via cross-attention. The query is derived solely from the last frame, aggregating contextual information from previous frames, and is then fed into the SMPL head to regress the single-frame body model parameters. Based on the overlapping window design, single-frame outputs from each window can be concatenated to reconstruct the full sequence with temporal continuity. It is then supervised with standard frame-level HMR losses, including 3D keypoints, 2D keypoints, SMPL parameters, and 3D vertices.
\begin{equation}
\mathcal{L}_{f} = \lambda_1 \mathcal{L}_{2D} + \lambda_2 \mathcal{L}_{3D} + \lambda_3 \mathcal{L}_{\text{SMPL}} + \lambda_4 \mathcal{L}_V
\end{equation}
\noindent\textbf{\textit{Inter-window temporal modeling.}} Leveraging the advantages of sliding window architecture, we add velocity regularization terms to penalize local joint velocity and acceleration relative to pelvis position on the continuous sequence of sliding window output, which are defined by:
\begin{equation}
\mathcal{L}_{v} = \lambda_5
\frac{
\sum_{i,t} c_{i,t} \, \left\| \mathbf{p}_{i,t} - \mathbf{p}_{i,t-1} \right\|_2^2
}{
\sum_{i,t} c_{i,t} + \epsilon
}
\end{equation}
\begin{equation}
\mathcal{L}_{a} = \lambda_6\frac{
\sum_{j,i} c_{j,i} \, \left\| \mathbf{p}_{j,i+1} - 2\mathbf{p}_{j,i} + \mathbf{p}_{j,i-1} \right\|_2^2
}{
\sum_{j,i} c_{j,i} + \epsilon
},
\end{equation}
where $\mathbf{p}$ is the joint position, $j$ denotes the joint index in the human skeleton, and $c$ is the confidence per joint provided in ground truth. The ultimate loss $\mathcal{L}$ is presented as:
\begin{equation}
\mathcal{L}=\mathcal{L}_f+\mathcal{L}_v+\mathcal{L}_a.
\end{equation}

\noindent\textbf{KV Cache Inference.}
The training process is conducted in a non-online fashion to ensure efficient parallelization. During inference, however, streaming inputs must be processed in a fully causal manner, avoiding the use of future observations for the current estimation. To this end, we introduce a key-value cache mechanism that stores the extracted features from previous frames before the attention correspondence matrix is computed. As illustrated in Fig.~\ref{fig_pipeline}, given the query $\mathbf{q}_i$ from the current frame and the cached keys $\mathbf{k}_{i-1}...\mathbf{k}_{i-N+1}$ and values $\mathbf{v}_{i-1}...\mathbf{v}_{i-N+1}$ from preceding frames, the attention operation is formulated as
\begin{equation}
\mathbf{A}_{\text{self}} = \mathrm{Softmax}\left(\frac{\mathbf{q}_i\mathbf{k}_i^\top}{\sqrt{d}}\right)\mathbf{v}_i,
\end{equation}
\begin{equation}
\mathbf{A}_{\text{cross}} = \mathrm{Softmax}\left(\frac{\mathbf{q}_i\mathbf{k}_{\text{prev}}^\top}{\sqrt{d}}\right)\mathbf{v}_{\text{prev}},
\end{equation}
where $d$ is the feature dimension, $\mathbf{k}_{\text{prev}}$ and $\mathbf{v}_{\text{prev}}$ are stacked from cache. This operation fuses spatial information among patches within the current frame and temporal information propagated from previous frames. In this way, we obtain temporally smooth human motion estimations in the camera coordinate. Then we need to estimate the camera trajectory to anchor the local motions into a shared global space.

\subsection{Human Centric Incremental SLAM}
\label{sec:method3.3}

Simultaneous Localization and Mapping (SLAM) has been extensively developed for robotic navigation, evolving from early heuristic-based approaches~\cite{nuchter20076dslam, sprickerhof2011heuristic, kim2015active} to recent learning-based methods~\cite{zhang2023goslam, neuralrecon, MASt3r-slam}. The online nature is embedded in these systems, enabling the robots to perceive the environment and plan their actions with minimal delay. Unlike prior work~\cite{tram, shen2024gvhmr, wham} using global optimized results, we take the incremental camera poses from SLAM, and combine them with per-frame human mesh. However, human-centric video does not fit the design principle of SLAM, since the static feature matching between views is violated by the large portion of human region with deformable textures and motion dynamics. Given that, we rethink the correlation between human and camera, and implement incremental physical constraints.

\noindent \textbf{Human $\rightarrow$ Camera.} 
One classical approach to deal with dynamic regions in SLAM is image space masking, typically using zero value, i.e., black pixels to fill the semantics segmented dynamic objects. Following this approach, we leverage SAM2~\cite{sam2} to segment the human region of each input frame. As show in Figure~\ref{fig_pipeline}, the SLAM system contains a frontend tracking and a backend graph-based optimization. Descriptors extracted from the previous frame need to find the best match in the following frame. Implementing masking at the image space forces the human regions to have no feature to be extracted, no matter using a heuristic or a learning based encoder. However, masking the human region itself is not sufficient, since the hard boundary between human mask and background pixels can also be encoded as unwanted features. To minimize its effect, we add a \textit{Dilation} and a \textit{Gaussian blur} on top of the human segment $C_i^h$, obtaining a soft confidence mask $C_i^{\text{soft}}$.
\begin{equation}
    C_i^{\text{soft}} = 
    \frac{
        G_{\sigma} * \big( C_i^h \oplus S_k^{(n)} \big)
    }{
        \displaystyle \max_{p} \big( G_{\sigma} * (  C_i^h \oplus S_k^{(n)} ) \big)
    }.
    \label{eq:softmask}
\end{equation}
where $S_k^{(n)}$ is the dilation kernel. In this way, we filter out most of the incorrect clues. The masked images with a static background are passed to SLAM to build up the camera trajectory and world geometry.

\noindent \textbf{Camera $\rightarrow$ Human.} Human instances moving in world in a continuous manner. It has to obay the physical rules such as no sudden jumps of position and no frequent change of acceleration or force. Previous work has explored to add physical constraints to HMR by introducing motion prior~\cite{ye2023slahmr, tokenhmr2024dwivedi}, but this cannot be generalized to scenarios where human is attaching to other transportation or performing out-of-distribution motions. Considering that the world-grounded human translation is dependent on per-frame camera translation in our framework, we impose smoothness constraints on the camera trajectory to indirectly regularize the temporal consistency of human motion. Exponential moving average (EMA) correction can be used to correct online estimation based on weighted history information. We apply this technique to our incremental camera trajectory from SLAM to suppress high-frequency jitters. 
For each frame $i$, we maintain a history buffer of size $B$ and define exponentially decaying weights:
\begin{equation}
    w_m = (1 - \alpha)^{B-1-m}, \quad m = 0, 1, \dots, B-1,
\end{equation}
which are normalized such that $\sum_m w_m = 1$. The history-weighted translation and the translation update are:
\begin{equation}
    \bar{\mathbf{t}}_i = \sum_{m=0}^{B-1} w_m\,\mathbf{t}_{i-m}, \quad
    \Delta \mathbf{t}_i = \mathbf{t}_i - \bar{\mathbf{t}}_i.
\end{equation}
We then estimate the average velocity magnitude $\overline{v}$ from recent frames. 
A velocity-dependent clamping threshold $\tau = \lambda_{\mathrm{clamp}}\overline{v}$ is used to avoid large jumps, i.e., if $\|\Delta \mathbf{t}_i\| > \tau$, the update is scaled to satisfy $\|\Delta \mathbf{t}_i\| = \tau$.
The final smoothed translation is given by:
\begin{equation}
    \mathbf{t}_i' = \bar{\mathbf{t}}_i + \alpha\,\Delta \mathbf{t}_i.
\end{equation}

For rotation smoothing, we use an EMA over quaternions $\mathbf{q}_i = (q_{ix}, q_{iy}, q_{iz}, q_{iw})$.
Before interpolation, we flip the quaternion hemisphere if necessary to ensure $\langle \mathbf{q}_{i-1}, \mathbf{q}_i \rangle > 0$, 
then apply a linear interpolation approximating SLERP:
\begin{equation}
    \mathbf{q}_i' = 
    \mathrm{normalize}\big((1 - \alpha)\mathbf{q}_{i-1}' + \alpha\,\mathbf{q}_i\big).
\end{equation}

The SLAM backend is also updated with these corrected translation and rotation for later optimization. The overall effect of EMA correction is further discussed in Sec.~\ref{sec:abla}.

To convert the arbitrary-scale SLAM outputs\footnote{Further discussion of SLAM output scale is providedd in Suppl.} into a metric world coordinate system, we employ the state-of-the-art metric depth estimation model MoGe-V2~\cite{mogev2} to predict per-frame metric depth, and calculate the scaler $s$ with SLAM depth map.
Notably, we ignore pixels corresponding to the human region in the scaler calculation. Since the human region is blurry in the visualized SLAM depth, and for in-the-wild human-centric videos, there are frequently appearing dolly zoom effects. Imagine the photographer is walking against the human, while zooming in the camera. The FOV is getting smaller, but the human-occupied scale in the image is not changing. In other words, the actual metric depth is getting larger, but not reflected by human region pixels. We keep the scaler consistent across the sequence, then multiply the SLAM output camera translation by this scaler to obtain the world coordinate camera pose.


\subsection{Frequency Domain Metric}
\label{sec:method3.4} 
In human motion estimation, different motion patterns exhibit varying temporal frequencies. A well-designed system should primarily capture low-frequency components, as natural human motion typically lies under 10Hz~\cite{freq1, freq2}. 
A common evaluation criterion for estimated human motion is the presence of jittering effects. Prior works typically quantify jitter using either \textit{dependent metrics} like Accel to compare the velocity change with ground truth, or \textit{independent metrics} like Jitter for acceleration change. However, these measures are highly sensitive to action types and often fail to capture the visually perceived temporal instability. To address this, we propose an alternative jitter representation based on spectrogram analysis.

Inspired by the Mel-Cepstral Distortion (MCD)~\cite{MCD} metrics in signal processing, we apply the short-time Fourier transform (STFT) to segment motion sequences into overlapping temporal windows and transform each segment into the frequency domain. This preserves both temporal and frequency information, yielding a time-varying frequency distribution that reflects motion stability over time.
Given a motion sequence 
$\mathbf{y}(i) \in \mathbb{R}^{F \times 3J}$, 
where $F$ is the frame number and $J$ is the number of joints. The spectrogram is computed as
\begin{equation}
    \mathbf{S}(i, f) = 
    \left| 
    \sum_{k=0}^{L-1} 
    \mathbf{y}(k) \,
    w(k - i) \,
    e^{-j 2 \pi f k / N_w}
    \right|,
    \label{eq:spectrogram}
\end{equation}
where $w(\cdot)$ is a Hann window function of length $N_{w} = n_{\text{fft}}$, 
and $L$ denotes the local temporal window size.
The magnitude $\left|\mathbf{S}(i, f)\right|$ represents the amplitude of each frequency component at frame index $i$. More implementation details, spectrogram visualization, and numerical quantification of jittering based on spectrogram are provided in the Suppl.

\begin{table}[t]
\centering
\begin{minipage}[t]{0.48\textwidth}
\centering
\caption{Comparison of camera coordinates HMR results using different models on 3DPW and EMDB-1 dataset. All metrics are in mm.} 
\normalsize
\setlength{\tabcolsep}{0.7mm}
\resizebox{\textwidth}{!}{
\begin{tabular}{p{2.8cm}|cccccccc}
\toprule
\multirow{2}{*}{Models} & \multicolumn{4}{c|}{3DPW} & \multicolumn{4}{c}{EMDB-1} \\
 & PA-MPJPE $\downarrow$ & MPJPE $\downarrow$ & PVE $\downarrow$ & Accel $\downarrow$ & PA-MPJPE $\downarrow$ & MPJPE $\downarrow$ & PVE $\downarrow$ & Accel $\downarrow$ \\
\midrule
\multicolumn{9}{l}{\textit{Per-Frame}} \\
HybrIK~\cite{hybrik2021Li} & 41.8 & 68.0 & 82.3 & --   & 65.6 & 103.0 & 122.2 & --   \\
HMR2.0~\cite{hmr2} & 44.4 & 69.9 & 82.2 & 18.1 & 60.7 & 98.3  & 120.8 & 19.9 \\
ReFit~\cite{refit} & 40.5 & 65.3 & 75.1 & 18.5 & 58.6 & 88.0  & 104.5 & 20.7 \\
\midrule
\multicolumn{9}{l}{\textit{\myBlue{Video-Offline}}} \\
TCMR~\cite{choi2021tcmr} & 52.7 & 86.5 & 101.4 &  6.0 & 79.8 & 127.7 & 150.2 &  5.3 \\
GLoT~\cite{shen2023glot} & 50.6 & 81.7 &  96.4 &  6.0 & 80.7 & 119.0 & 140.8 &  5.4 \\
GLAMR~\cite{glamr} & 51.1 & --   & --  &  8.0 & 73.8 & 113.8 & 134.9 & 33.0 \\
WHAM~\cite{wham} & 37.5 & 59.8 & 71.5 & 6.6 & 52.0 & 81.6 &  96.9 & 5.3 \\
TRAM~\cite{tram} & 35.6 & 63.9 & 69.6 &  \textbf{\myBlue{4.9}} & 45.7 & 74.4 & 86.6 & \textbf{\myBlue{4.9}} \\
GVHMR~\cite{shen2024gvhmr} & 37.0 & \textbf{\myBlue{56.6}} & 68.7 & -- & 44.5 & 74.2 & 85.9 & -- \\
PHMR~\cite{wang2025prompthmr} & \textbf{\myBlue{35.5}} & 56.9 & \textbf{\myBlue{67.3}} & -- & \textbf{\myBlue{40.1}} & \textbf{\myBlue{68.1}} & \textbf{\myBlue{79.2}} & --\\
\midrule
\multicolumn{9}{l}{\textit{\myGreen{Video-Online}}} \\
TRACE~\cite{trace} & 50.9 & 79.1 &  95.4 & 28.6 & 71.5 & 110.0 & 129.6 & 25.5 \\
Human3R~\cite{chen2025human3r} & 44.1 & 71.2 & 84.9 & -- & 48.5 & \textbf{\myGreen{73.9}} & \textbf{\myGreen{86.0}} & -- \\
OnlineHMR (Ours) & \textbf{\myGreen{43.7}} & \textbf{\myGreen{69.9}} & \textbf{\myGreen{83.7}} & \textbf{\myGreen{6.4}} & \textbf{\myGreen{46.0}} & 74.0 & 86.1 & \textbf{\myGreen{9.0}} \\
\bottomrule
\end{tabular}}
\label{tab:cameracoord}
\end{minipage}

\begin{minipage}[t]{0.48\textwidth}
\centering
\caption{\small Evaluation of human global trajectory and motions on EMDB-2 dataset. 
RTE is in \%, ERVE is in \textit{mm}/frame, and the other pose metrics are in \textit{mm}.}
\small
\setlength{\tabcolsep}{0.5mm}
\resizebox{\textwidth}{!}{
\begin{tabular}{l|ccccccc}
\toprule
Models & PA-MPJPE $\downarrow$ & WA-MPJPE$_{100}$  $\downarrow$ & W-MPJPE$_{100}$ $\downarrow$ & RTE $\downarrow$ & ERVE $\downarrow$ \\
\midrule
\multicolumn{6}{l}{\textit{\myBlue{Video-Offline}}} \\
GLAMR~\cite{glamr}  & 56.0 & 280.8 & 726.6  & 11.4 & 18.0 \\
SLAHMR~\cite{ye2023slahmr}  & 61.5 & 326.9 & 776.1 & 10.2 & 19.7 \\
WHAM (w/DPVO)~\cite{wham}  & 38.2 & 135.6 & 354.8  & 6.0 & 14.7 \\
COIN~\cite{coin} & \textbf{\myBlue{32.7}} & 152.8 & 407.3 & 3.5 & -- \\
TRAM~\cite{tram}  & 38.1 & 76.4 & 222.4 & 1.4 & \textbf{\myBlue{10.3}} \\
JOSH~\cite{josh} & -- & \textbf{\myBlue{68.9}} & \textbf{\myBlue{174.7}} & \textbf{\myBlue{1.3}} & -- \\
GVHMR (w/DPVO) ~\cite{shen2024gvhmr}  &  -- & 111.0 & 276.5 & 2.0 & -- \\
PHMR~\cite{wang2025prompthmr} & -- & 71.0 & 216.5 & \textbf{\myBlue{1.3}} & -- \\
\midrule
\multicolumn{6}{l}{\textit{\myGreen{Video-Online}}} \\
TRACE~\cite{trace} & 58.0 & 529.0 & 1702.3 & 17.7 & 370.7 \\
JOSH3R~\cite{josh} & -- & 220.0 & 661.7 & 13.1 & -- \\
Human3R~\cite{chen2025human3r}  & -- & 112.2 & \textbf{\myGreen{267.9}} & \textbf{\myGreen{2.2}} & -- \\
OnlineHMR  (Ours) & \textbf{\myGreen{40.1}} & \textbf{\myGreen{93.5}} & 310.4 & \textbf{\myGreen{2.2}} & \textbf{\myGreen{12.4}} \\
\bottomrule
\end{tabular}}
\label{tab:worldcoord}
\vspace{-4mm}
\end{minipage}
\end{table}
\section{Experiment}
\subsection{Datasets and Metrics}

\begin{figure*}[t]
    \vspace{-2mm}
    \centering    \includegraphics[width=1.0\linewidth]{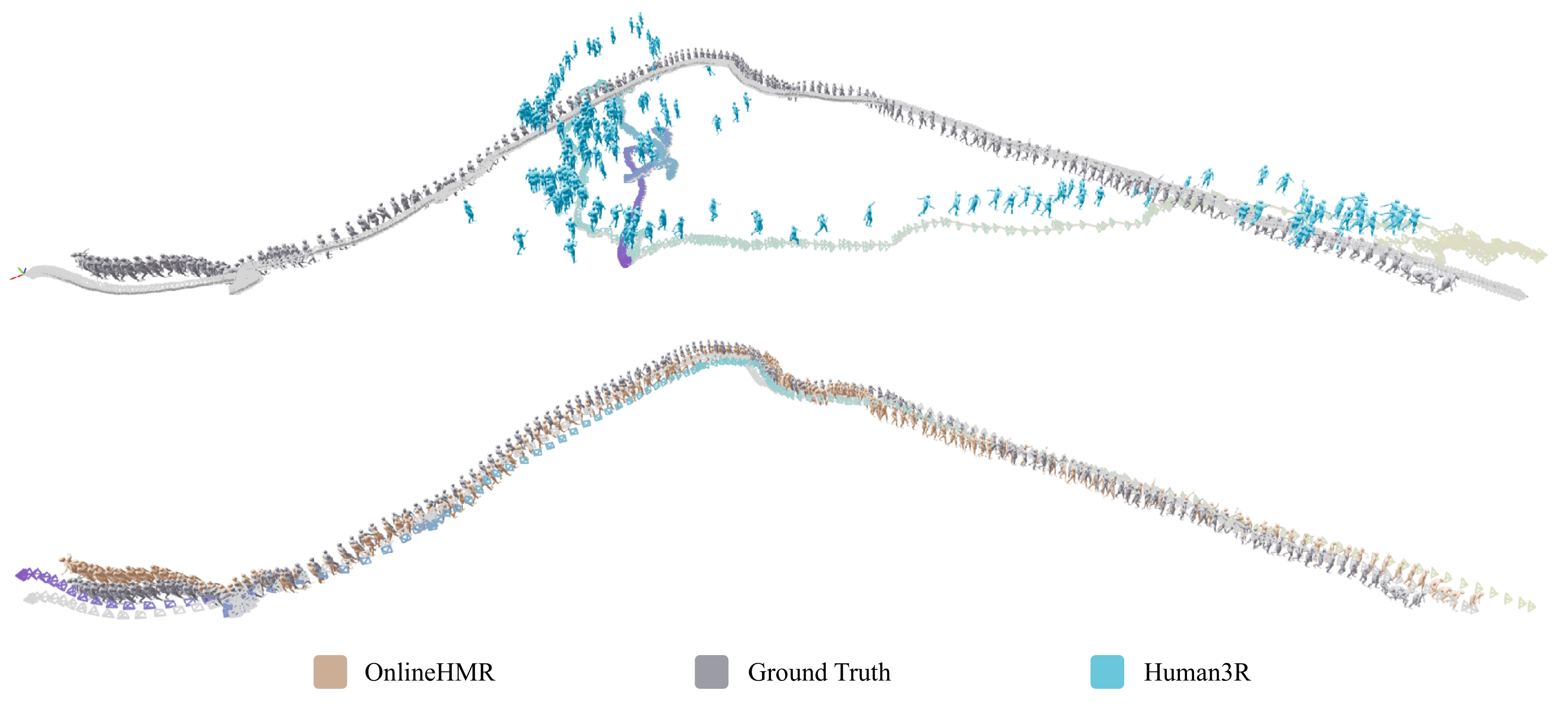}
    \vspace{-2mm}
   \caption{Quantitative comparison of OnlineHMR and Human3R on the same EMDB-2 video with ground truth after world coordinate alignments. }\label{fig:supplemdb224}
   \vspace{-4mm}
   \label{fig_world_overlap_compare_2}
\end{figure*}

Following TRAM~\cite{tram}, we use 3DPW~\cite{3dpw} and EMDB-1~\cite{emdb} for camera coordinates, and EMDB-2~\cite{emdb} for world coordinates evaluation. Camera coordinates metrics include mean per-joint position error (MPJPE), Procrustes-aligned per-joint position error (PA-MPJPE), per-vertex error (PVE), and acceleration error (Accel), measured in millimeters. For world coordinates evaluation, we follow prior work, dividing each sequence into 100-frame segments and evaluating 3D joint errors using W-MPJPE, which aligns the first two frames, and WA-MPJPE, which aligns the entire segment. We use the root translation error (RTE, in \%) after rigid alignment, normalized by the total displacement, to evaluate trajectory accuracy over long sequences and egocentric-frame root velocity error (ERVE) to measure the root motion accuracy.

\begin{figure}[h]
    \centering    \includegraphics[width=1.0\linewidth]{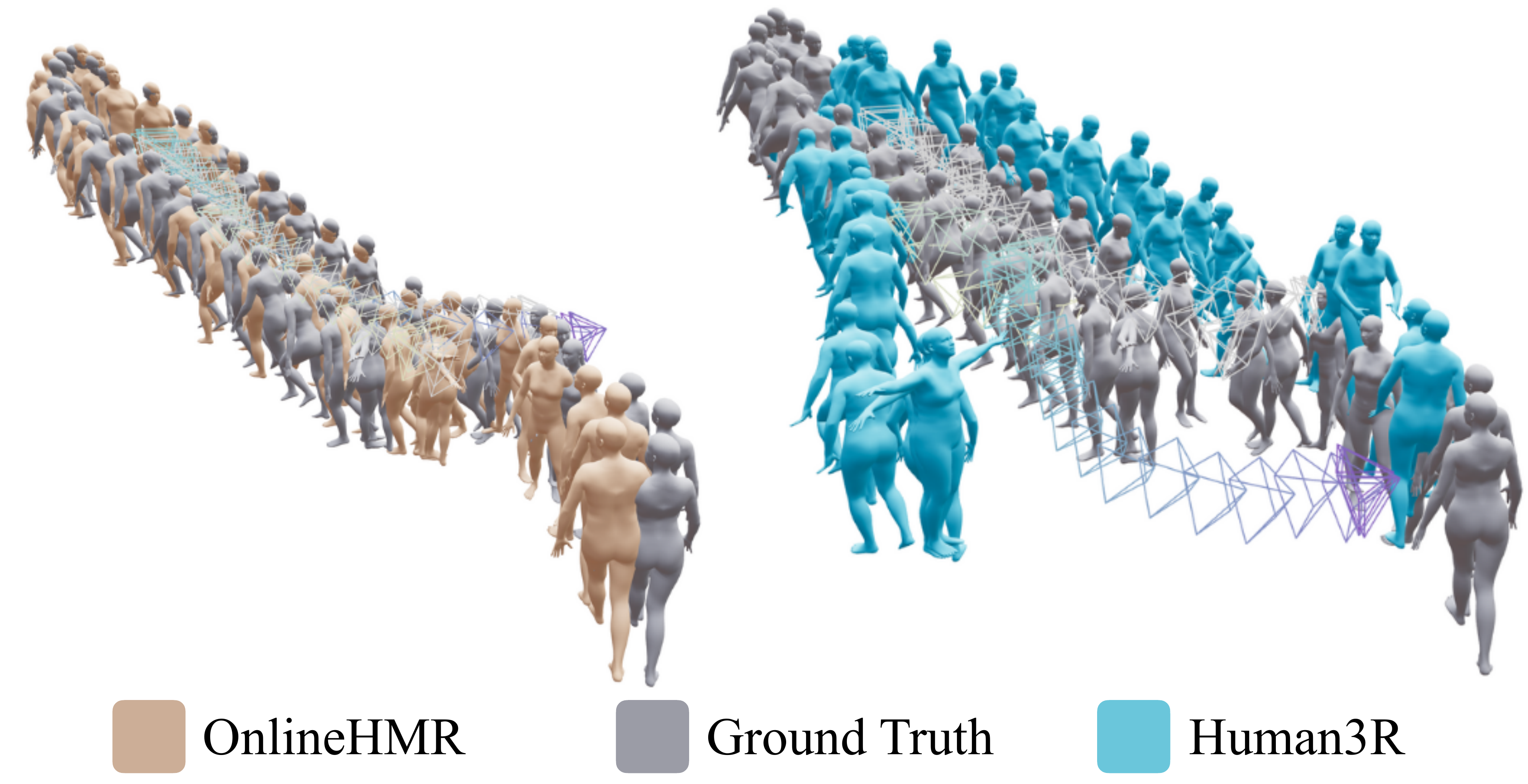}
   \caption{Quantitative comparison of OnlineHMR and Human3R on the same EMDB-2 video with ground truth after world coordinate alignments. }\label{fig:supplemdb277}
   \vspace{-5mm}
   \label{fig_world_overlap_compare_1}
\end{figure}
We use BEDLAM~\cite{bedlam}, 3DPW~\cite{3dpw}, and H3.6M~\cite{human36m} to train the model, which converges around 52K iterations. The training is conducted on one Nvidia 80GB H100 GPU. More details are in Suppl.


\subsection{Results Analysis}
\label{sec:exp4.2}
\noindent\textbf{Faithfulness \& Consistency.} The camera-coordinate results reflect the quality of the recovered local motion. As shown in Tab.~\ref{tab:cameracoord}, our method achieves the best performance in both accuracy and temporal consistency among all \textit{Video-Online} approaches. Moreover, it exhibits no significant degradation compared to chunk-based offline methods, despite the constraint of operating in an online setting. Although prior work WHAM~\cite{wham} is online in camera coordinates, its VO component performs sliding window local optimization, and outputs the results with the previous frame being optimized by later observations. Thus, we categorize it into the offline group. 

\begin{figure}[t]
    \centering    
    \includegraphics[width=1.0\linewidth]{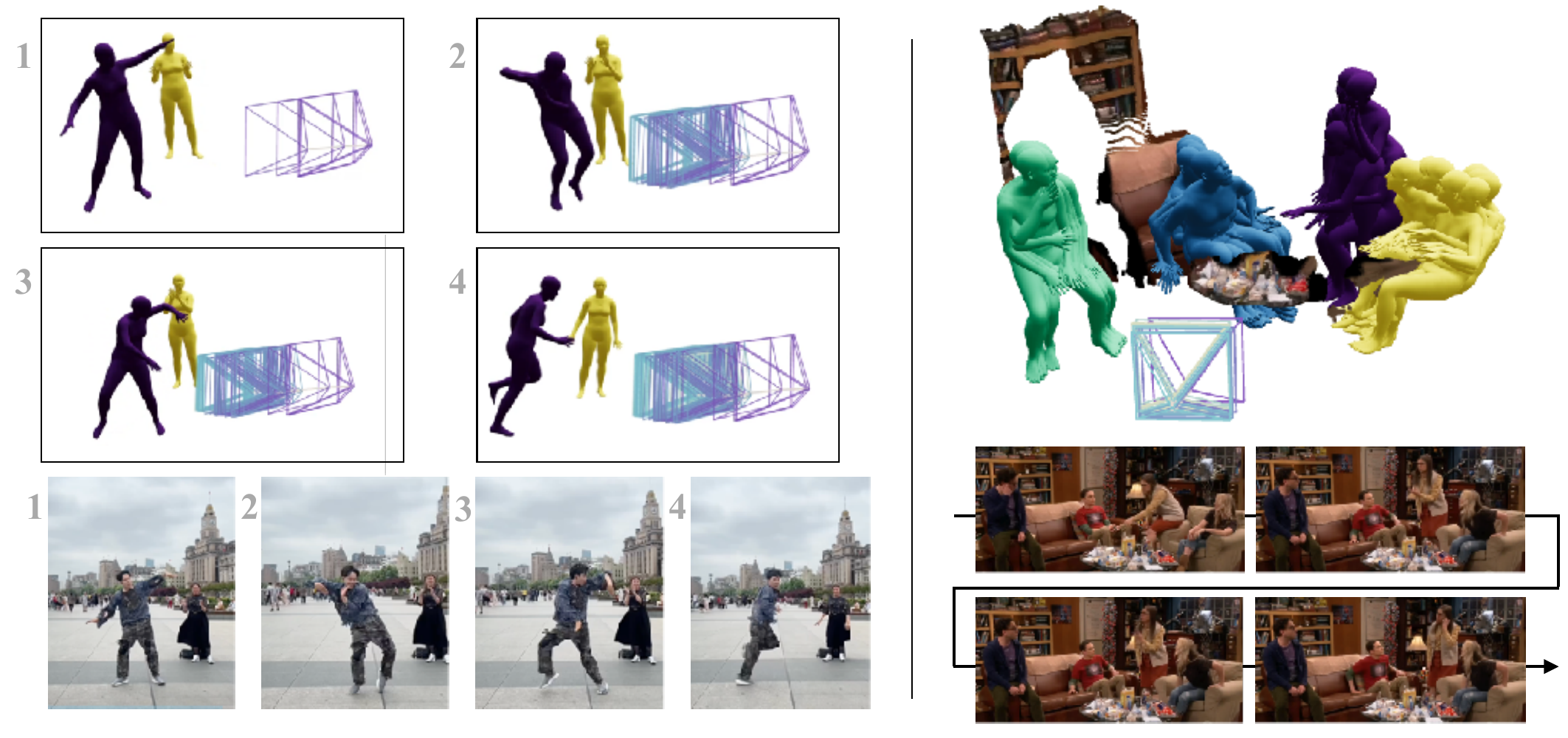}
    \vspace{-4mm}
   \caption{Visualization of multi-individual, diverse scene cases. More examples are in Suppl.}
   \label{fig:multidiverse}
   \vspace{-5mm}
\end{figure}

For the world coordinate result, as shown in Tab.~\ref{tab:worldcoord} we have leading performance on most metrics. Note that the W-MPJPE is lagging behind, mostly due to the inaccuracy introduced by metric scale conversion. Since the main difference between WA-MPJPE and W-MPJPE is the number of frames used to align the predicted results and the ground truth sequence, a good performance in WA-MPJPE with a degradation in W-MPJPE indicates that although the position and orientation of the world origin point have been aligned, the scale of later incremental estimation is not well-estimated, but after scaling the whole sequence, the global shape of trajectory and human motion are both well reconstructed. 
We visualize the predicted results by overlapping them with the ground truth and provide a comparison to concurrent work, Human3R, in Fig.~\ref{fig_world_overlap_compare_2} and Fig.~\ref{fig_world_overlap_compare_1}. Additionally, our method is able to generalize to multi-individual settings, with visualization in Fig.~\ref{fig:multidiverse} and dynamic results on our webpage.

\noindent\textbf{Efficiency.} One main advantage of the online system is its efficiency and minimal delay. We report two metrics, frame-per-second (FPS) and average delay time per frame (Avg. Delay), with corresponding analysis shown in Tab.~\ref{tab:supplfps}.
\label{sup:efficiency}

\begin{table}[h]
\centering
\vspace{-4mm}
\caption{Efficiency comparison of world coordinate HMR methods. The average delay time is presented in seconds.}

\resizebox{0.48\textwidth}{!}{
\begin{tabular}{l|c|c|c|c}
\toprule
Models & Online & FPS$\uparrow$ & Avg. Delay (s)$\downarrow$ & WA-MPJPE$_{100}$ $\downarrow$\\
\midrule
TRACE~\cite{trace} & \cmark & 24.6 & 0.04 & 529.0 \\
GLAMR~\cite{glamr}  & \xmark & 17.2 & 14.16 & 280.8\\
SLAHMR~\cite{ye2023slahmr}  & \xmark & 0.1 & 2435 & 326.9 \\
WHAM (w/ DPVO)~\cite{wham}  & \xmark & 9.3 & 26.18 & 135.6 \\
TRAM ~\cite{tram}  & \xmark & 2.1 & 115.95 & 76.4\\
GVHMR (w/ DPVO)~\cite{shen2024gvhmr} & \xmark & 12.1 & 20.12 & 111.0 \\
Human3R~\cite{chen2025human3r} & \cmark & 4.8 & 0.21 & 112.2 \\
OnlineHMR (Ours) & \cmark & 3.3 & 0.30 & 93.5\\
\bottomrule
\end{tabular}}
\label{tab:supplfps}
\vspace{-5mm}
\end{table}

\noindent\textbf{\textit{FPS}}. We test the FPS on a single RTX 6000 Ada GPU, using the same video across all open-sourced world coordinates HMR methods. For methods that have multi-stage processing, we record the total run time and calculate the average FPS. 
Note that many factors, including server occupancy, video length, and video content, may influence FPS. Since video content decides the descriptors, and the different quality of descriptors and the complexity of the trajectory make the backend matching take various times. Also, the speed of the incremental SLAM method set the upper bound of FPS. In addition, for methods using global optimization, the total length of the video affects the efficiency of optimization. We also want to point out that offline methods support to batchify frames into chunks, and running the inference in parallel if the input does not come in a streaming way, while the online method only allows processing one frame each time to support a streaming process. Therefore, we suggest the reader treat this \textit{FPS} term only as a sub-important reference.

\noindent\textbf{\textit{Avg. Delay.}} Another metric we want to present is the average delay time per frame. It shows the average waiting time of all frames before yielding the result of this frame from network estimation. Here, the input video is assumed to be streaming. Note that the Avg. Delay metric is linearly correlated to video length $F$ for offline methods, since:
\begin{equation}
\text{total delay} = \frac{F(F-1)}{2\times \text{FPS}},
\end{equation}
\begin{equation}
\text{average delay} = \frac{\text{total delay}}{F} = \frac{F-1}{2\times \text{FPS}}.
\end{equation}
Results in Tab.~\ref{tab:supplfps} show that online methods have significantly lower delay compared to offline methods. Although our OnlineHMR has the max Avg. Delay among all online methods, we have a better accuracy in both camera coordinate and world coordinate results.

\section{Ablation Study}
\label{sec:abla}

We conduct an ablation study regarding design choices that are mentioned in Sec.~\ref{sec:method3.2} and Sec.~\ref{sec:method3.3}. For additional ablations, please refer to Suppl.

\noindent\textbf{Velocity Regularization.}
We compare the results on 3DPW and EMDB-1 w/ or w/o velocity regularization. In Tab.~\ref{tab:ablajitter}, both the \textit{dependent} Accel and \textit{independent} Jitter metrics perform better with the regularization, indicating an improved consistency with the ground truth and a less frequent acceleration changing.
\begin{table}[h]
\centering
\caption{The comparison of jittering effect w/ or w/o the velocity regularization.}
\resizebox{0.46\textwidth}{!}{
\begin{tabular}{l|cc|cc}
\toprule
\multirow{2}{*}{Models} & \multicolumn{2}{c|}{3DPW} & \multicolumn{2}{c}{EMDB-1} \\
& Accel $\downarrow$ & Jitter $\downarrow$ & Accel $\downarrow$ & Jitter$\downarrow$ \\
\midrule
w/o Velocity Regularization & 8.9 & 32.3 & 15.7 & 70.1 \\
w/ Velocity Regularization & 6.4 & 19.5 & 9.0 & 33.7 \\
\bottomrule
\end{tabular}}
\label{tab:ablajitter}
\vspace{-2mm}
\end{table}

\begin{figure}[h]
    \centering    \includegraphics[width=0.9\linewidth]{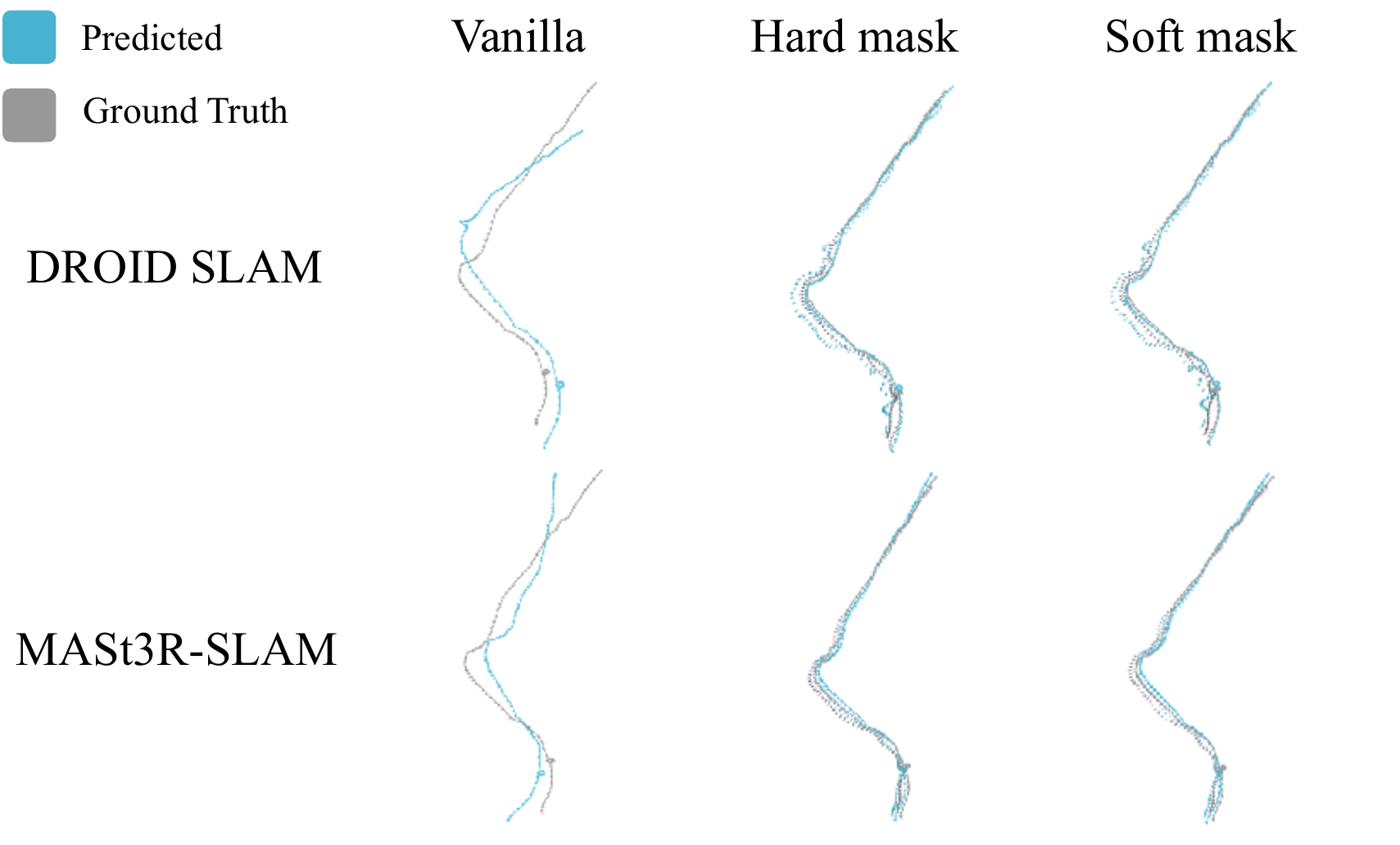}
    \caption{Visualization of estimated camera trajectory compared with GT.}
    \vspace{-6mm}
    \label{fig_abcamtraj}
\end{figure}

\noindent\textbf{Soft Human Mask.} We apply either a hard or soft mask to the human region in the SLAM input images and evaluate their impact on camera trajectory estimation. The Absolute Trajectory Error (ATE) measures the alignment quality between the predicted and ground-truth trajectories after a similarity transformation. The result is in Tab.~\ref{tab:slam_mask_comparison}. We also present the world coordinates human reconstruction metrics that are affected by the camera trajectory estimation results. Applying a soft mask on top of MASt3R-SLAM achieves the most accurate camera trajectory, as shown in Fig.~\ref{fig_abcamtraj}, effectively mitigating incorrect feature matching caused by dynamic human regions.

\begin{table}[h]
\centering
\vspace{-2mm}
\caption{Comparison of ATE using different masking strategies for DROID-SLAM and MAST3R-SLAM. Lower is better.}
\resizebox{0.39\textwidth}{!}{
\begin{tabular}{lccc}
\toprule
Method & Vanilla & Hard Mask & Soft Mask \\
\midrule
DROID-SLAM & 2.52 & 1.55 & 1.07 \\
MAST3R-SLAM & 1.22 & 0.96 & 0.83\\
\bottomrule
\end{tabular}}
\label{tab:slam_mask_comparison}
\end{table}
\vspace{-6mm}

\section{Conclusion}

We presented OnlineHMR, a fully online framework for world-coordinate human mesh recovery. Avoiding using future information for current estimation, we curate the camera coordinates online HMR branch, effectively fuse information from nearby frames, while maintaining local temporal consistency. We also rethink the role of the camera and human in streaming input, then apply constraints to the incremental SLAM system to yield a more faithful camera trajectory reconstruction, which indirectly benefits the global naturalness of human motion.
The full framework satisfies core requirements of online processing, including causality, faithfulness, temporal consistency, and efficiency, while remains competitive with chunk-based offline models.

\noindent \textbf{Limitations:} As OnlineHMR relies on incremental SLAM and history-based EMA correction, it assumes continuous viewpoints and struggles with abrupt camera switches or multi-camera inputs. Future work includes improving robustness under severe viewpoint discontinuities and exploring multi-camera setting extensions.

\section{Acknowledgement}
We thank Shubham Tulsiani and Michael Kaess for their insightful feedback, Aniket Agarwal for early-stage brainstorming, Taru Rustagi, Ananya Bal, Joel Julin, Kallol Saha, Hongwen Zhang, Zihan Wang, Jiatong Shi for helpful discussion, Nathalie Chang, Rena Ju for early-stage survey, and the anonymous reviewers for suggestions.

{
    \small
    \bibliographystyle{ieeenat_fullname}
    \bibliography{main}
}

\newpage
\clearpage
\setcounter{page}{1}
\maketitlesupplementary

\section{Overview}

The supplementary material includes the subsequent components.

\begin{itemize}
    \item \textbf{Relative Concept Explanation}
    \begin{itemize}[label=-]
        \item Explanation of the up-to-scale estimation of the SLAM system (Sec.~\ref{sup:slam} -- Sec. 3.3).
        \item Implementation details and numerical quantification regarding spectrogram analysis (Sec.~\ref{sup:frequencymetric} -- Sec. 3.4).
    \end{itemize}
    \item \textbf{Supplementary Experiments and Analysis.}
    \begin{itemize}[label=-]
        \item Details of the main experiment (Sec.~\ref{sup:exp} -- Sec. 3.2 \& Sec. 4.1).
        \item Ablation on sliding window size shown by camera coordinate human reconstruction metrics (Sec.~\ref{supplabla sliding window size} -- Sec. 5). 
        \item Ablation on different masking strategies for world HMR precision. (Sec.~\ref{sup:mask} -- Sec. 5).
    \end{itemize}
    \item \textbf{Additional Visualization}
    \begin{itemize}[label=-]
        \item World HMR visualization on custom videos (Sec.~\ref{sup:adviscustom} -- Sec. 4.2).
        \item World HMR visualization on multi-individual and diverse scene cases (Sec.~\ref{sup:viswscene} -- Sec. 4.2).
        \item Visualization on ablation results (Sec.~\ref{sup:spec} -- Sec. 5).
        \item Visualization on failure cases. (Sec.~\ref{sup:visfailurecases} -- Sec. 6).
    \end{itemize}
    \item \textbf{Supplementary Video (on website)} 
\end{itemize}

\section{Up-to-Scale SLAM System}
\label{sup:slam}
In Sec. 3.3, we adopt MASt3R-SLAM~\cite{MASt3r-slam} to estimate the world-coordinate camera trajectory and mentioned its up-to-scale reconstruction of camera transformation. 
  
Built upon the two-view 3D reconstruction model MASt3R~\cite{MASt3R}, MASt3R-SLAM~\cite{MASt3r-slam} is among the first to integrate strong 3D priors into an incremental SLAM framework, achieving fast inference of up to 15 FPS on a single RTX 4090 GPU. Although MASt3R~\cite{MASt3R} is trained with a metric regression loss, where the normalization factor used in its predecessor DUSt3R~\cite{wang2024DUSt3R} is removed to enable metric-scale reconstruction, MASt3R-SLAM still produces up-to-scale results. 

This is because MASt3R-SLAM does not explicitly constrain the incremental camera pose as $\mathbf{T}_{cw}\in \mathbf{SE}(3)$ during training.
Instead, it defines all camera poses as $\mathbf{T}_{cw}\in \mathbf{Sim}(3)$, allowing an additional scale component. Consequently, while the scale of the estimated scene and camera trajectory are temporally consistent, they remain up-to-scale rather than metrically scaled.
We estimate the scaler $s$ from metric depth. In practice, we find that the scaler is around 1 in most of the testset sequences, indicating an inheritated metric scale reconstruction ability from MASt3R~\cite{MASt3R}. But for texture-less and highly dynamic sequences, the scaler factor drifts away from 1 and plays an important role in metric scale recovery.

\section{Details about Frequency Domain Metrics}
\label{sup:frequencymetric}
In Section 3.4, we propose a frequency-domain representation of jittering effects. The mathematical formulation and implementation details are as follows.

As mentioned in the main paper, we obtain the motion spectrogram using the Short Time Fourier Transform (STFT) to preserve both the time-axis and frequency-axis. The reason for directly flattening the 3D joint position to 1D is that we care about jittering among all dimensions of the space and all joints. So there is no subsample of joints or separation of xyz axes. 

We use a Hann window with length $N_{w} = n_{\text{fft}}=128$, hop length $L=32$. After transforming the flattened signal using STFT, we apply $\text{abs}(\cdot)$ to get the magnitude, and interpolate it to the original sequence length. After these operations, the spectrogram $\left|\mathbf{S}(i, f)\right|$ has its y-axis representing frequency bins, and x-axis denoting temporal frame index. 

\begin{figure*}[h]
    \centering    \includegraphics[width=1.0\linewidth]{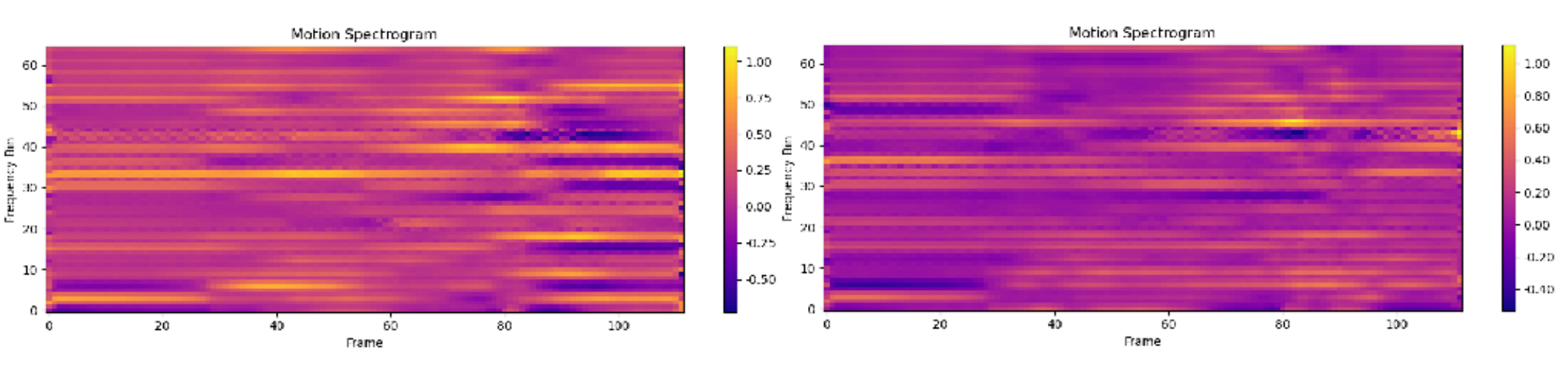}
    \caption{Difference spectrograms computed as GT-Pred, visualizing discrepancies in the time–frequency domain. The left figure corresponds to a model trained for 1K iterations, and the right for 52K iterations. Lower values (darker/closer to zero) indicate better alignment with the ground truth. The converged model (52K) exhibits substantially reduced differences.}
    \vspace{-3mm}
    \label{fig:abonSpec}
\end{figure*}

Following prior work on signal processing~\cite{MCD}, we calculate the root mean square error (RMSE) and correlation (Corr) based on the spectrogram statistics, and derive metrics based on that:
\[
\text{MSE}
= 
\frac{1}{N}
\sum_{i,f}
\left(
S^{\text{gt}}_{i,f} - S^{\text{pred}}_{i,f}
\right)^2,
\]
\[
\text{RMSE} = \sqrt{\text{MSE}},
\qquad
\text{RMSE}_{\text{norm}} = 
100 \times\frac{\text{RMSE}}{\sigma_{\text{gt}} + \epsilon}.
\]
where \(\sigma_{\text{gt}}\) is the standard deviation of \(S^{\text{gt}}_{i,f}\).
For Corr, we first calculate its mean:
\[
\mu_{\text{gt}} = \frac{1}{N}\sum_{i,f} S^{\text{gt}}_{i,f},
\qquad
\mu_{\text{pred}} = \frac{1}{N}\sum_{i,f} S^{\text{pred}}_{i,f}.
\]
Then the correlation coefficient is: 
\[
\text{Corr} =
\frac{
\sum_{i,f}
\left(S^{\text{gt}}_{i,f}-\mu_{\text{gt}}\right)
\left(S^{\text{pred}}_{i,f}-\mu_{\text{pred}}\right)
}{
\sqrt{
\sum_{i,f}\left(S^{\text{gt}}_{i,f}-\mu_{\text{gt}}\right)^2
\;
\sum_{i,f}\left(S^{\text{pred}}_{i,f}-\mu_{\text{pred}}\right)^2
}
}\]
\[
\text{Corr}_{\text{norm}} = 100 \times \frac{1 - \text{Corr}}{2}.
\]
We use the  $\text{RMSE}_{\text{norm}}$ and the $\text{Corr}_{\text{norm}}$ as dependent metrics and show an example of our OnlineHMR checkpoints trained by 1K iters and 52K iters. Both metrics are the lower the better, as shown in Fig.~\ref{fig:abonSpec}.

\begin{table}[h]
\centering
\caption{RMSE$\downarrow$ / Corr $\downarrow$ results.}
\scriptsize
\begin{tabular}{l|ccc}
\hline
Dataset & GVHMR & TRAM & Ours \\
\hline
3DPW & 3.59 / 0.02 & 17.01 / 0.64 & 19.91 / 0.82\\
EMDB-1 & 75.52 / 0.12 & 25.82 / 1.24 & 24.98 / 1.41 \\
\hline
\end{tabular}
\label{tab_freqcompare}
\vspace{-4mm}
\end{table}

We report camera coordinate results on recent methods and ours, as shown in Tab.~\ref{tab_freqcompare}. Ours (focus online) is comparable to the offline baseline TRAM in frequency amplitude (RMSE) and pattern (Corr) distribution similarity w.r.t GT. Interestingly, GVHMR shows better pattern reconstruction but slightly larger deviation in amplitude on EMDB-1.

\section{Details about the Main Experiment}
\label{sup:exp}
\noindent\textbf{Loss functions.} The standard per-frame HMR loss function in Sec.3.2 includes 3D keypoints, 2D keypoints, SMPL parameters, and 3D vertices components, shown as.
\begin{equation}
\mathcal{L}_{f} = \lambda_1 \mathcal{L}_{2D} + \lambda_2 \mathcal{L}_{3D} + \lambda_3 \mathcal{L}_{\text{SMPL}} + \lambda_4 \mathcal{L}_V
\label{eq:losses}
\end{equation}
We denote $\mathbf{J}_{3D}$ as the integration of 3D position for all the joints $\{\mathbf{p}_1, \mathbf{p}_2, ..., \mathbf{p}_{j},...,\mathbf{p}_{J}\}$. Similarly, $\boldsymbol{\Theta}$ indicates the SMPL pose parameters, $\boldsymbol{V}$ denotes 3D vertices for all the joints. Then each component in ~\eqref{eq:losses} is:
\begin{align}
\mathcal{L}_{2D} &= \left\| \hat{\mathbf{J}}_{2D} - \Pi(\mathbf{J}_{3D}) \right\|_F^2, \\
\mathcal{L}_{3D} &= \left\| \hat{\mathbf{J}}_{3D} - \mathbf{J}_{3D} \right\|_F^2, \\
\mathcal{L}_{\text{SMPL}} &= \left\| \hat{\boldsymbol{\Theta}} - \boldsymbol{\Theta} \right\|_2^2, \\
\mathcal{L}_V &= \left\| \hat{\mathbf{V}} - \mathbf{V} \right\|_F^2,
\end{align}
where $\hat{}$ indicates the ground truth, F is the total frame number, and $\Pi$ is the projection function. The loss weights $\lambda_1=5.0$, $\lambda_2=5.0$, $\lambda_3=1.0$, $\lambda_4=1.0$. And the velocity regularization weights mentioned in the main paper are $\lambda_5=10.0$, $\lambda_6=5.0$.

\noindent\textbf{Parameters setup.} In training, we first process the input video feature sequence into 16-frame chunks, with training batch size=24, as used in TRAM~\cite{tram}. Then apply a window slicing using length $N=3$, $stepsize=1$. Within each window, we estimate the result of frame $N-1$ only, while taking frame $N-3$ and $N-2$ as conditions, which we name as \textbf{\textit{intra-window information fusion}}. This is supervised by frame-level HMR losses. Then, for \textbf{\textit{inter-window temporal modeling}}, we stack the output of all windows together into a 14-frame chunk, and add additional supervision of velocity regularizations. The model is trained on an Nvidia 80GB H100 GPU.

\section{More Ablation on Sliding Window Size.}
\label{supplabla sliding window size}
We ablate on the sliding window size 3-6 as reported in Tab.~\ref{tab:sws}. Results show the accuracy reach optimal at window size=4, but Accel increases along with the window size getting larger. We infer this is due to the temporal sensitivity of Accel Metric. Larger windows fuse more past frames, introducing a delay effect that amplifies the acceleration deviations from GT.

\begin{table}[h]
\centering
\caption{Ablation study on sliding window size (SWS).}
\vspace{-1em}
\renewcommand{\arraystretch}{0.9}
\scriptsize
\setlength{\tabcolsep}{1.2pt}
\begin{tabular}{lcccccccc}
\hline
& \multicolumn{4}{c}{3DPW} & \multicolumn{4}{c}{EMDB-1} \\
\hline
SWS & PA-MPJPE$\downarrow$ &MPJPE$\downarrow$  & PVE$\downarrow$ & 
 ACCEL$\downarrow$ & PA-MPJPE$\downarrow$ &MPJPE$\downarrow$  & PVE$\downarrow$ & 
 ACCEL$\downarrow$\\ \hline
3 & 43.7 & 69.9 & 83.7 & \textbf{6.4} & \textbf{46.0} & 74.0 & 86.1 & \textbf{9.0} \\
4 & \textbf{41.7} & \textbf{64.9} & \textbf{78.6} & \textbf{6.4} & 46.8 & \textbf{73.8} & 85.3 & 9.1  \\
5 & 43.2 & 70.8 & 85.7 & 6.6 & 46.7 & 74.5 & 86.5 & 9.4 \\
6 & 42.5 & 65.5 & 80.1 & 6.6 & 46.9 & 74.3 & \textbf{84.4} & 9.6 \\ 
\hline
\end{tabular}
\label{tab:sws}
\vspace{-4mm}
\end{table}

\section{More Ablation on Masking Strategy}
\vspace{-4mm}
\label{sup:mask}
\begin{table}[h]
\centering
\caption{Comparison of different masking strategies on world coordinate human reconstruction upon MASt3R-SLAM~\cite{MASt3r-slam}}
\resizebox{0.4\textwidth}{!}{
\begin{tabular}{l|c|c|c|c}
\toprule
Strategy & WA-MPJPE & W-MPJPE & RTE & ERVE \\
\midrule
Vanilla & 119.6 & 412.9 & 4.1 & 14.4\\
Hard Mask & 112.6 & 386.8 & 3.2 & 13.3\\
Soft Mask & 93.5 & 310.4 & 2.2 & 12.4\\
\bottomrule
\end{tabular}}
\label{tab:worldcompare}
\end{table}
In Sec. 5, we provided a comparison of the camera trajectory accuracy on different masking strategies. Here, we additionally present the world coordinates human reconstruction metrics that are also affected by the camera trajectory estimation results, as shown in Tab.~\ref{tab:worldcompare}.

\section{Additional Visualization on Custom Video}

\label{sup:adviscustom}

\begin{figure*}[h]
    \centering    \includegraphics[width=1.0\linewidth]{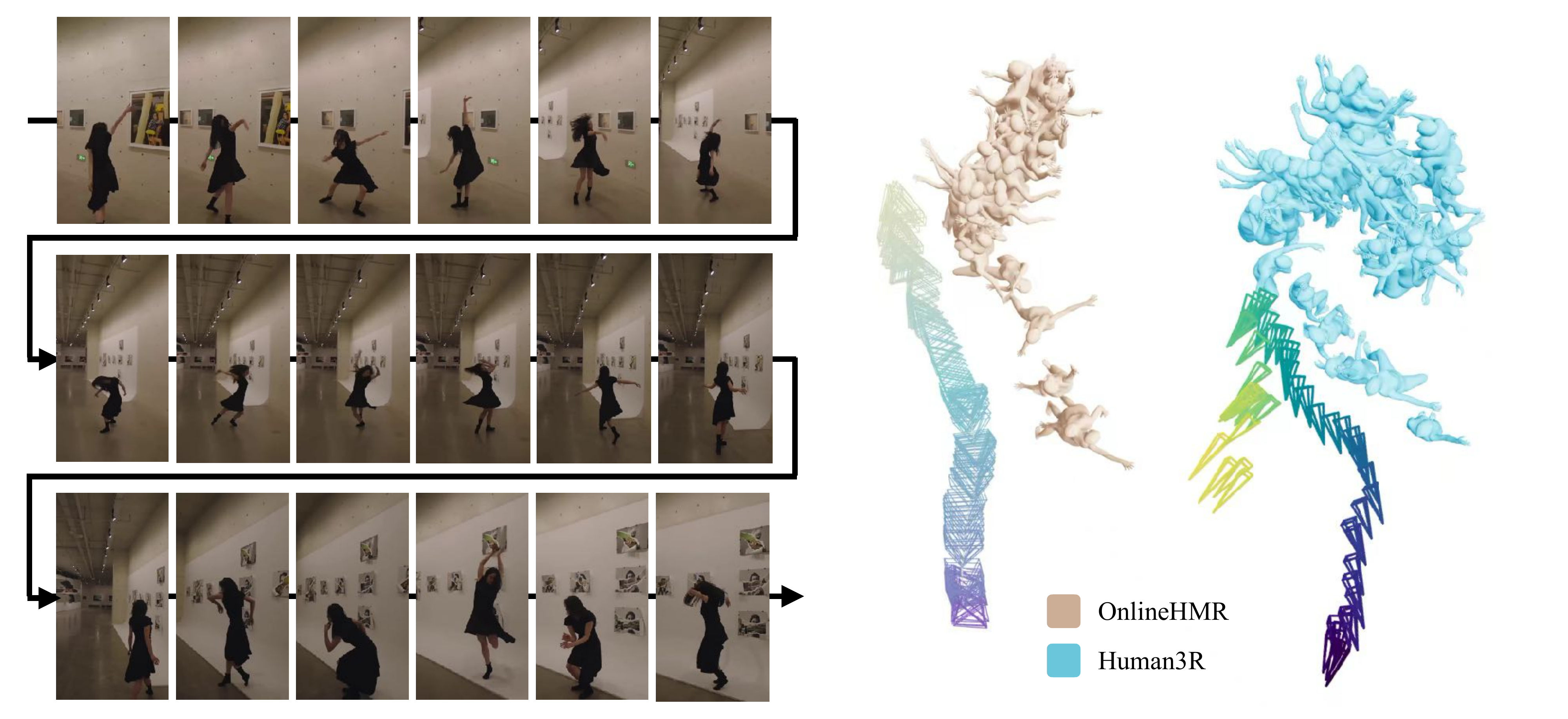}
   \caption{Quantitative comparison of OnlineHMR and Human3R on a custom video of a famous dancer. OnlineHMR produces a faithful reconstruction of the human trajectory, whereas Human3R yields trajectories that appear compressed and crowded together.}\label{fig:supplcustomvideo}
   \vspace{-5mm}
\end{figure*}

\begin{figure*}[h]
    \centering    \includegraphics[width=1.0\linewidth]{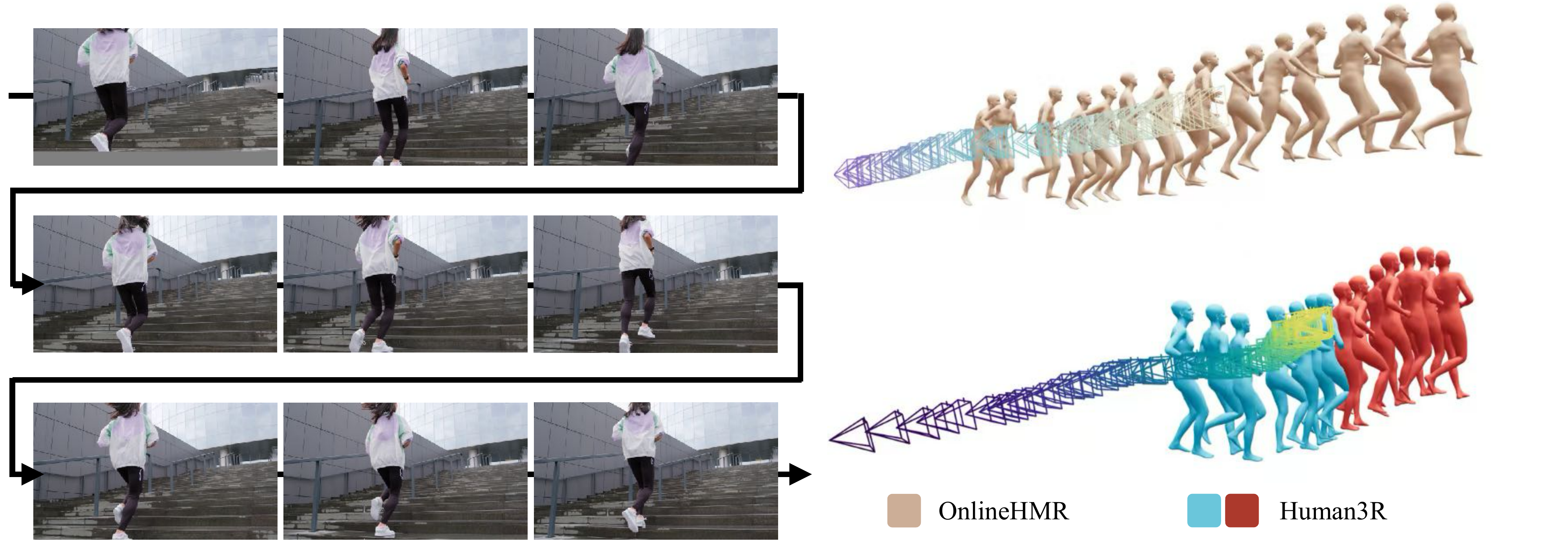}
   \caption{Quantitative comparison of OnlineHMR and Human3R on a custom video of running up stairs. }  \label{fig:supplcustomvideo2}
      \vspace{-4mm}
\end{figure*}

As shown by Fig.~\ref{fig:supplcustomvideo} and Fig.~\ref{fig:supplcustomvideo2}, our OnlineHMR reconstructs a more faithful world coordinate camera trajectory and human mesh compared to concurrent work Human3R~\cite{chen2025human3r}. Fig.~\ref{fig:supplcustomvideo2} additionally shows that Human3R sometimes loses tracking of the person (blue) in the video and mixes up with a second person (red).
Dynamic results are in the supplementary video.

\section{Visualization on Multi-Individual and Diverse Scene}
\label{sup:viswscene}
As stated in the main paper, our main focus is on world coordinate human mesh recovery. However, we can also obtain the scene point cloud from the SLAM part. We illustrate an example of a human mesh and camera trajectory with scene geometry.

\begin{figure}[h]
    \centering    \includegraphics[width=0.9\linewidth]{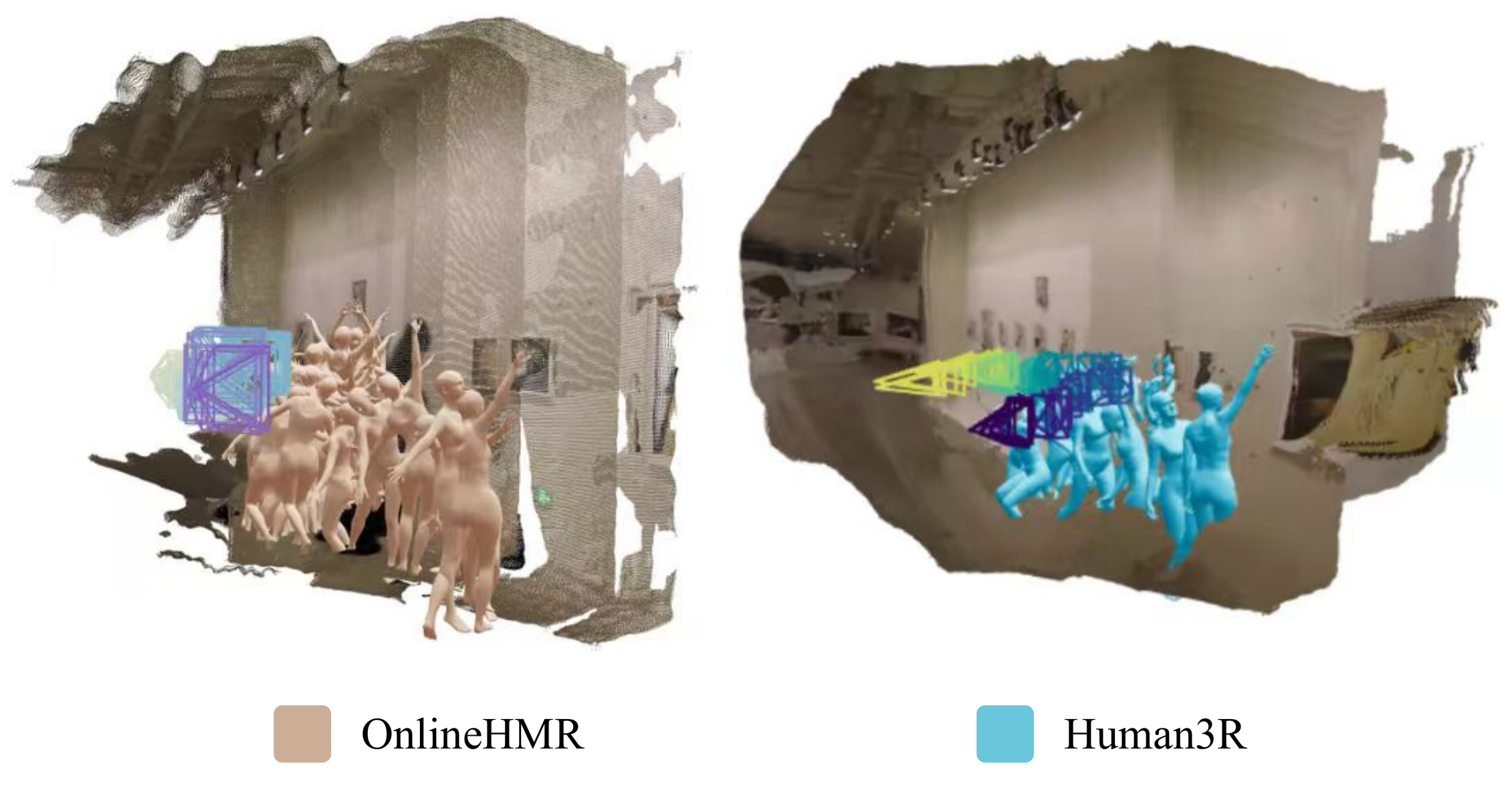}
    \caption{Using the same video input with Fig.~\ref{fig:supplcustomvideo}, we show the global human mesh, camera trajectory, and the final updated scene.}
    \label{fig:viswscene}
    \vspace{-4mm}
\end{figure}
\begin{figure*}[t]
    \centering    \includegraphics[width=0.9\linewidth]{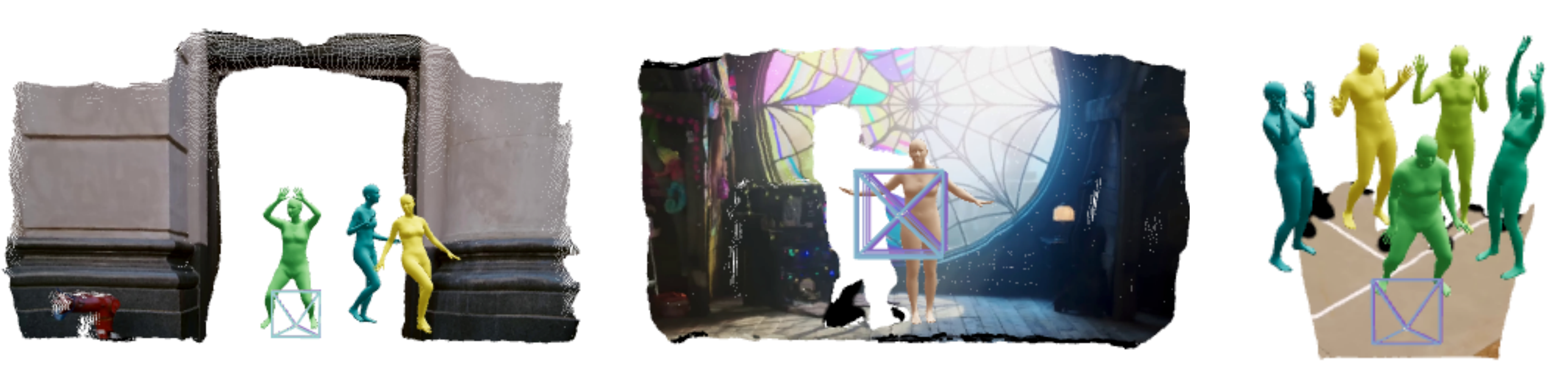}
    \caption{More examples with multiple individuals and a diverse scene. Dynamic results can be found on the demo page.}
    \label{fig:moreexsonsuppl}
\end{figure*}

Our method is also able to generalize to multi-individual and complex scene scenarios. Examples are shown in Fig.~\ref{fig:moreexsonsuppl}.

\section{Visualization on Ablation Results}
\label{sup:spec}
\begin{figure}[h]
    \centering    \includegraphics[width=0.9\linewidth]{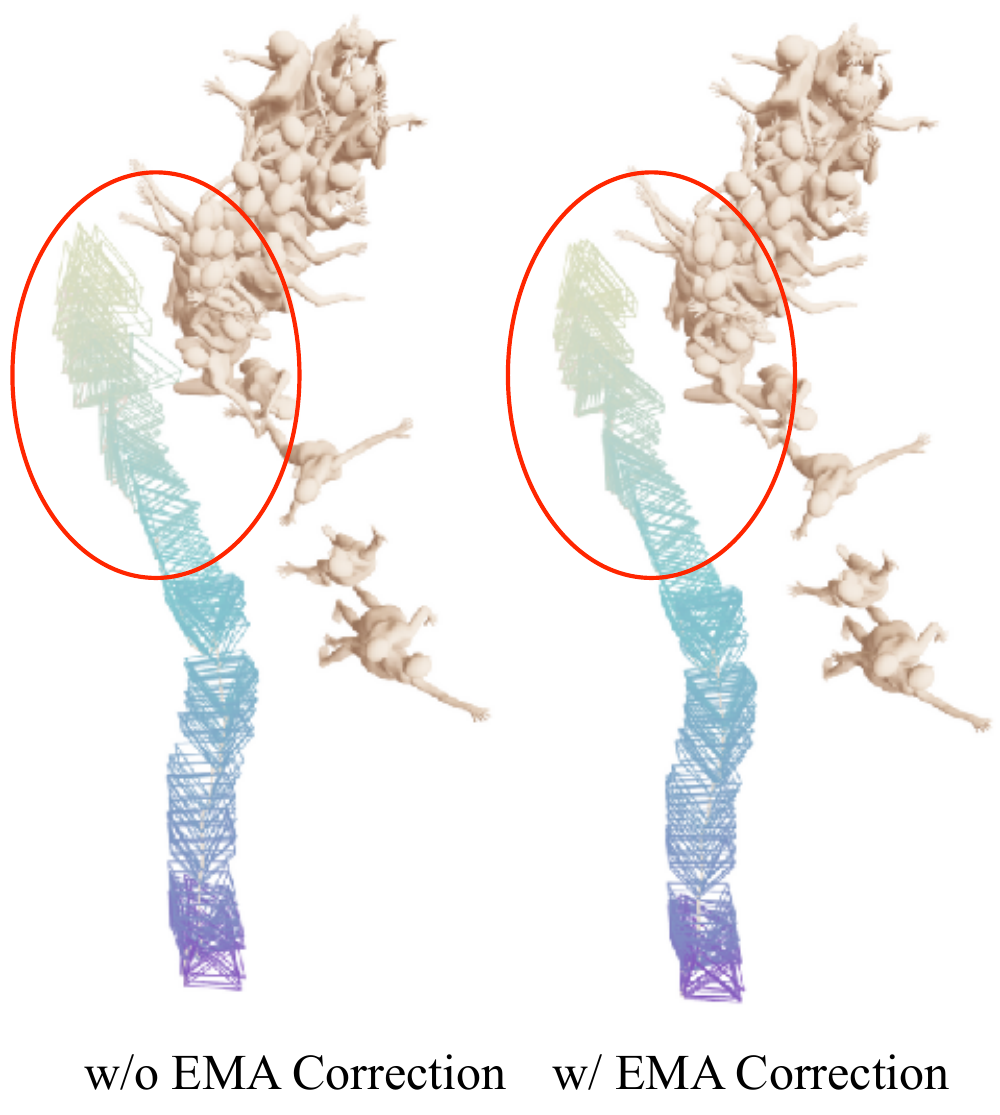}
    \vspace{-4mm}
    \caption{Qualitative results w/ and w/o EMA correction on custom videos.}
    \label{fig:abonEMAj}
    \vspace{-4mm}
\end{figure}

\noindent\textbf{EMA Correction.} As stated in Sec. 3.3, EMA correction on camera translation and rotation indirectly imposes a smoothness for world coordinate human motion. We show a qualitative comparison of custom video in Fig.~\ref{fig:abonEMAj}. The results demonstrate less jittering effect on the camera and world coordinate human translation with EMA correction.

\begin{figure}[h]
    \centering    \includegraphics[width=0.9\linewidth]{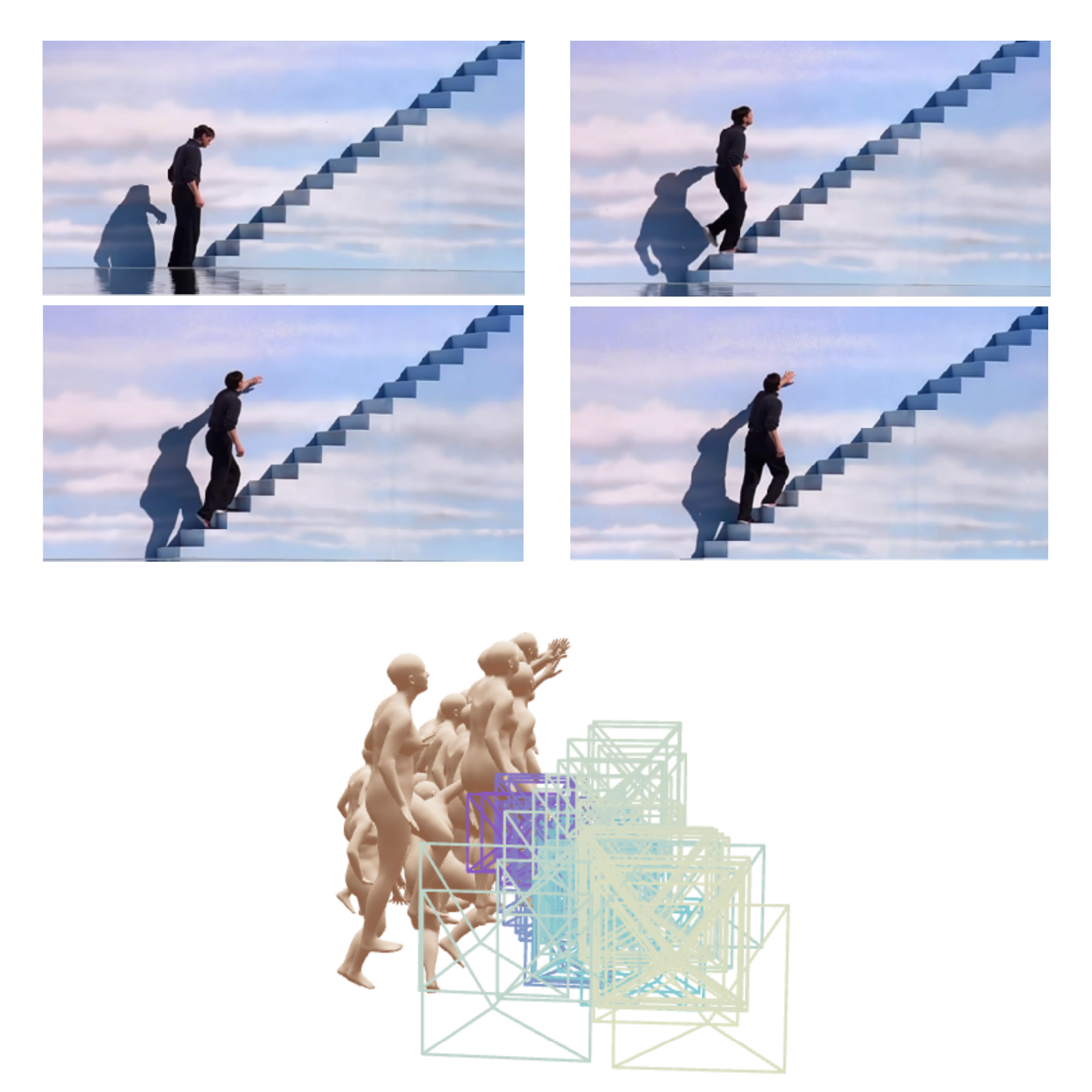}
    \caption{Example of failure cases.}
    \label{fig:visfailurecases}
        \vspace{-4mm}
\end{figure}

\section{Visualization on Failure Cases}
\label{sup:visfailurecases}
As shown in Fig.~\ref{fig:visfailurecases}, our method fails on inputs with repetitive textures and dynamic environments, such as shadow. Also, the current design is not suitable for camera switching cases.


\end{document}